\pdfoutput=1
\documentclass[11pt]{article}
\usepackage[preprint]{acl} 
\usepackage{times}
\usepackage{latexsym}
\usepackage[T1]{fontenc}
\usepackage[utf8]{inputenc}
\usepackage{microtype}
\usepackage{inconsolata}
\usepackage{graphicx}
\usepackage{subcaption} 
\usepackage{booktabs}   
\usepackage{wrapfig}
\usepackage{threeparttable}
\usepackage{amsmath}
\usepackage{amsfonts}
\usepackage{bm}
\usepackage{textcomp}
\usepackage{xcolor}
\usepackage{colortbl}
\usepackage{tcolorbox}
\usepackage{color}
\usepackage{enumitem}
\usepackage{multirow}
\usepackage{geometry}
\usepackage[ruled,vlined]{algorithm2e}
\usepackage{cleveref}
\crefformat{section}{\S#2#1#3}
\crefformat{subsection}{\S#2#1#3}
\crefformat{subsubsection}{\S#2#1#3}
\crefrangeformat{section}{\S\S#3#1#4 to~#5#2#6}
\crefmultiformat{section}{\S\S#2#1#3}{ and~#2#1#3}{, #2#1#3}{ and~#2#1#3}
\crefmultiformat{subsection}{\S\S#2#1#3}{ and~#2#1#3}{, #2#1#3}{ and~#2#1#3}
\Crefformat{figure}{#2Fig.~#1#3}
\Crefmultiformat{figure}{Figs.~#2#1#3}{ and~#2#1#3}{, #2#1#3}{ and~#2#1#3}
\Crefformat{table}{#2Tab.~#1#3}
\Crefmultiformat{table}{Tabs.~#2#1#3}{ and~#2#1#3}{, #2#1#3}{ and~#2#1#3}
\Crefformat{appendix}{Appx.#2#1#3}
\crefmultiformat{appendix}{Appx.#2#1#3}{ and~#2#1#3}{, #2#1#3}{ and~#2#1#3}
\crefformat{algorithm}{Alg.~#2#1#3}
\Crefformat{equation}{Eq.~#2#1#3}
\newcommand{\myparagraph}[1]{\vspace{0.3em}\noindent{{\bf #1.}}}
\newcommand{\modelname}{\textsc{S-VCO}\xspace}
\newcommand{\dataname}{\textsc{MVC}\xspace}

\definecolor{matchingColor}{HTML}{05B0F0}
\definecolor{contradictingColor}{HTML}{F49443}
\definecolor{noImageColor}{HTML}{7F7F7F}
\definecolor{baseColor}{HTML}{767171}
\definecolor{dpoColor}{HTML}{FF9300}
\definecolor{mdpoColor}{HTML}{0070C0}
\definecolor{svcoColor}{HTML}{7030A0}
\definecolor{goodShadeColor}{HTML}{D2FBE1}

\title{Symmetrical Visual Contrastive Optimization: Aligning Vision-Language Models with Minimal Contrastive Images}
\author{
Shengguang Wu,
Fan-Yun Sun,
Kaiyue Wen,
Nick Haber
\\
Stanford University
\\
\\
\url{https://s-vco.github.io/}
}

\begin{document}
\maketitle
\begin{abstract}
Recent studies have shown that Large Vision-Language Models (VLMs) tend to neglect image content and over-rely on language-model priors, resulting in errors in visually grounded tasks and hallucinations.
We hypothesize that this issue arises because existing VLMs are not explicitly trained to generate texts that are accurately grounded in fine-grained image details.
To enhance visual feedback during VLM training, we propose \modelname (\textbf{S}ymmetrical \textbf{V}isual \textbf{C}ontrastive \textbf{O}ptimization), a novel finetuning objective that steers the model toward capturing important visual details and aligning them with corresponding text tokens.
To further facilitate this detailed alignment, we introduce \dataname, a paired image-text dataset built by automatically filtering and augmenting visual counterfactual data to challenge the model with hard contrastive cases involving \textbf{M}inimal \textbf{V}isual \textbf{C}ontrasts.
Experiments show that our method consistently improves VLM performance across diverse benchmarks covering various abilities and domains, achieving up to a 22\% reduction in hallucinations, and significant gains in vision-centric and general tasks. Notably, these improvements become increasingly pronounced in benchmarks with higher visual dependency.
In short, \modelname offers a significant enhancement of VLM's visually-dependent task performance while retaining or even improving the model's general abilities.
\end{abstract}

\begin{figure}[t]
\centering
\includegraphics[width=1.0\linewidth]{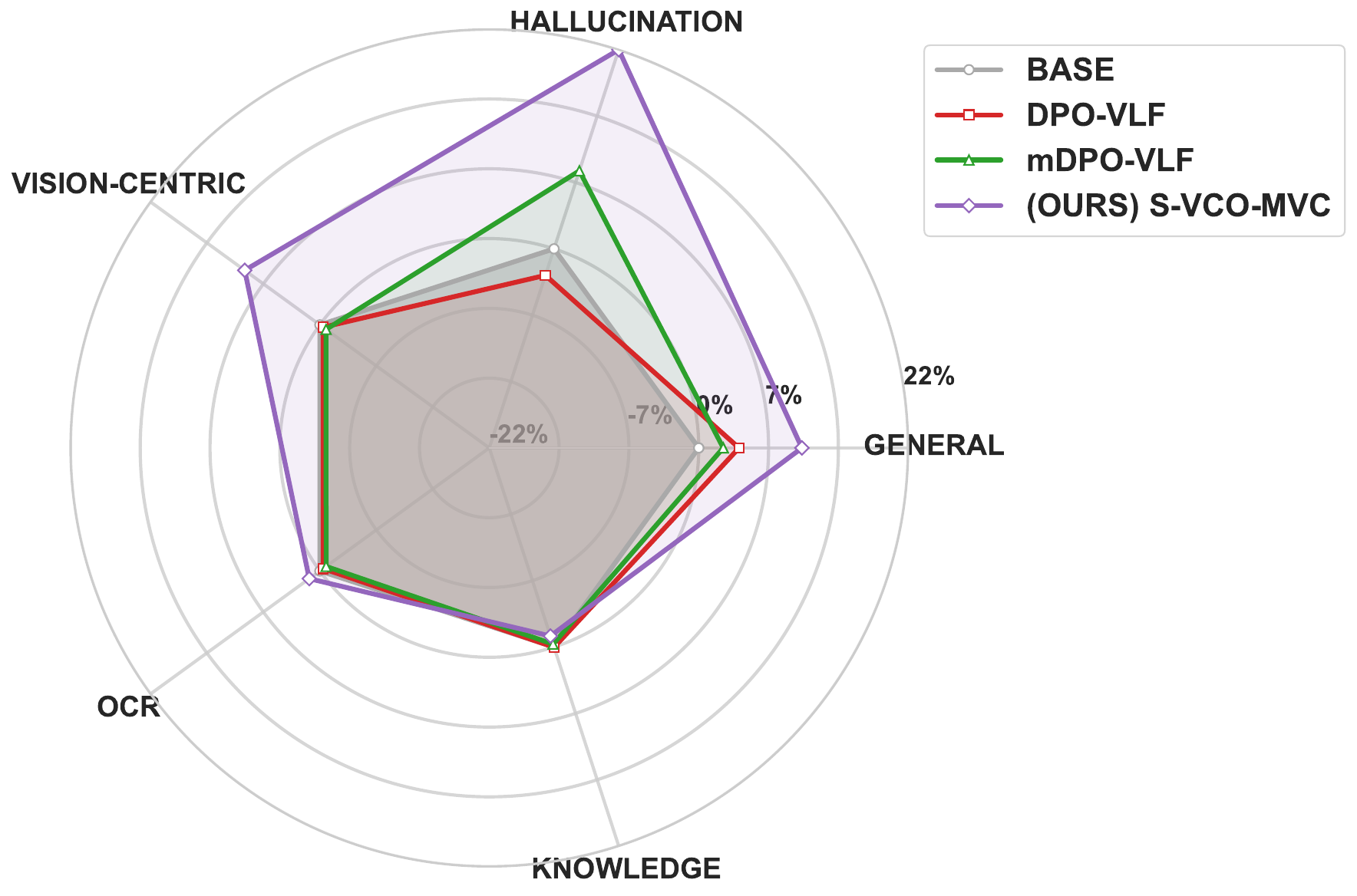}
\caption{\textbf{Improvement over the base-VLM grouped by the test ability domain of benchmarks.} Our \modelname delivers the most significant overall improvement across nearly all domains, with particularly strong gains in reducing visual hallucinations. In vision-centric and general capability domains, \modelname also achieves considerable performance boosts over the base-VLM, outperforming existing preference tuning methods including DPO and mDPO (discussed in more detail in \Cref{sec:experiment_results}).}
\label{fig:improve_per_domain}
\end{figure}

\begin{figure}[t]
\centering
\includegraphics[width=1.0\linewidth]{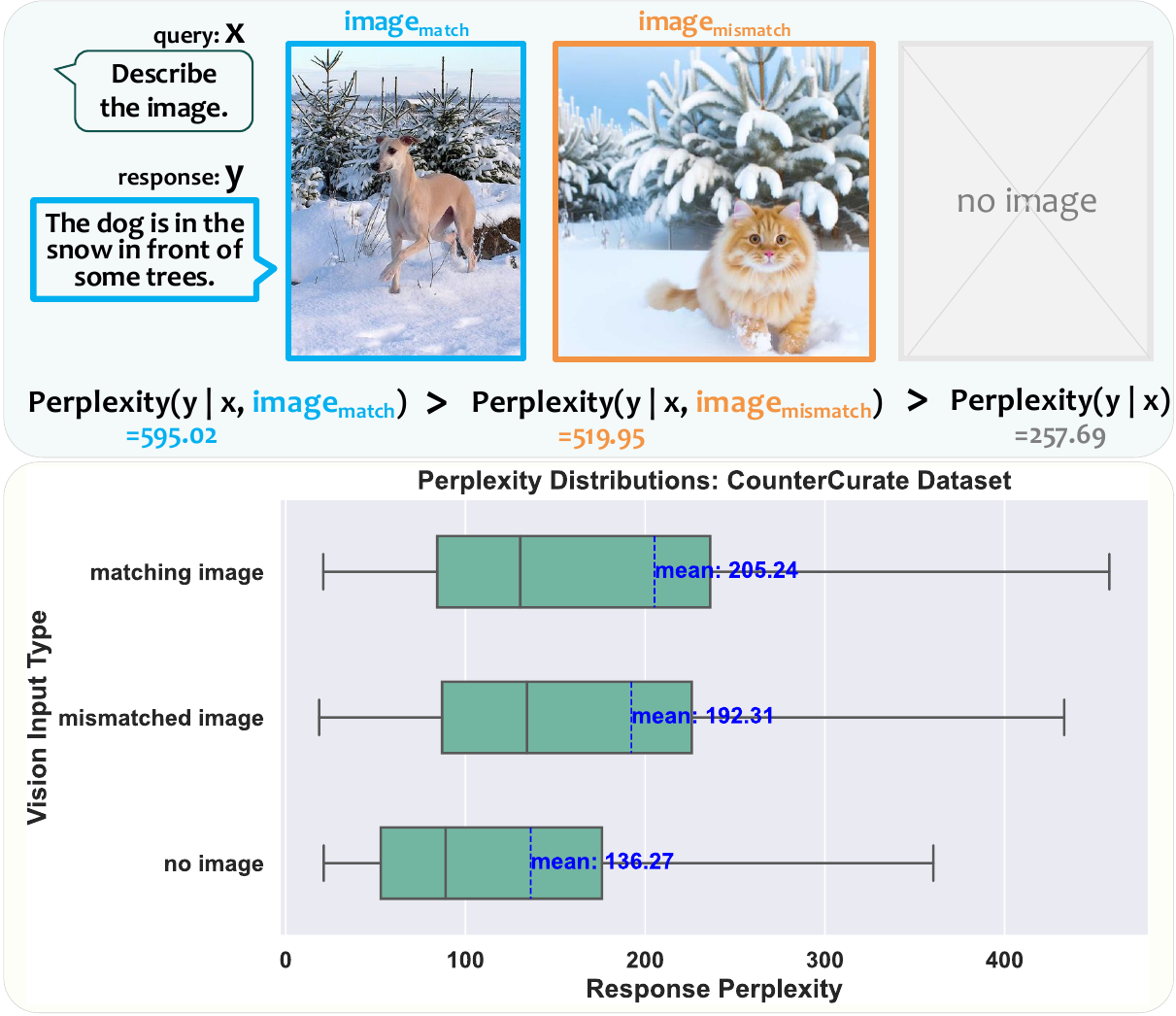}
\caption{\textbf{Upper Part}: A comparison of a VLM's perplexity (PPL) for generating a text caption when the input is \textcolor{matchingColor}{an image matching the text}, 
\textcolor{contradictingColor}{an image contradicting the text}, 
or \textcolor{noImageColor}{no image input at all}. Intuitively, the PPL should be lowest when the image matches the text. However, the current VLM exhibits the lowest PPL without any image input and the highest PPL given the matching image. \textbf{Lower Part}: This counterintuitive pattern holds across $1,000$ random examples with these visual counterfactual pairs extracted from CounterCurate dataset (\Cref{sec:data_visual_counterfactual_data}).}
\label{fig:visual_neglect}
\end{figure}

\section{Introduction} \label{sec:intro}
Large Vision-Language Models (VLMs) tend to over-rely on their language models, leading to neglect of visual content. This problem manifests as visual hallucinations across tasks like perception \citep{mmvp}, reasoning \citep{chen2024quantifying}, and in-context learning \citep{symdpo}. 
Studies like \citet{cambrian} show that VLMs exhibit only limited performance gains when vision inputs are enabled compared to having no vision inputs. This behavior is reflected in the metrics across many popular vision-language benchmarks \citep{sqa, mmmu, mathvista, ai2d, realworldqa, textvqa}.
Our own perplexity-based evaluations reveal similar patterns of \textbf{visual neglect}. As illustrated in \Cref{fig:visual_neglect}, we measured a base VLM's perplexity (PPL) when generating a caption matched to one image (\emph{e.g.}, ``image\textsubscript{match}'' with a ``dog'') while contradicting the other image (\emph{e.g.}, ``image\textsubscript{mismatch}'' with a ``cat''). We also tested the model's PPL without any vision input (``no image'').
Intuitively, the model should exhibit the lowest PPL when given the matching image, as the aligned visual information would make the caption easier to generate. 
However, results reveal the opposite pattern: the model's PPL lowest when no image input is provided at all, and highest when presented with the correct image (the ``dog'') -- higher than the mismatched image (the ``cat'') as condition.
Across $1,000$ random samples from CounterCurate \citep{countercurate} — a large collection of such paired counterfactual images — the average perplexity distribution follows this counterintuitive pattern. This highlights the model's tendency to disregard visual information even when they are critical for generating accurate texts.

\begin{figure*}[t]
\centering
\includegraphics[width=1.0\linewidth]{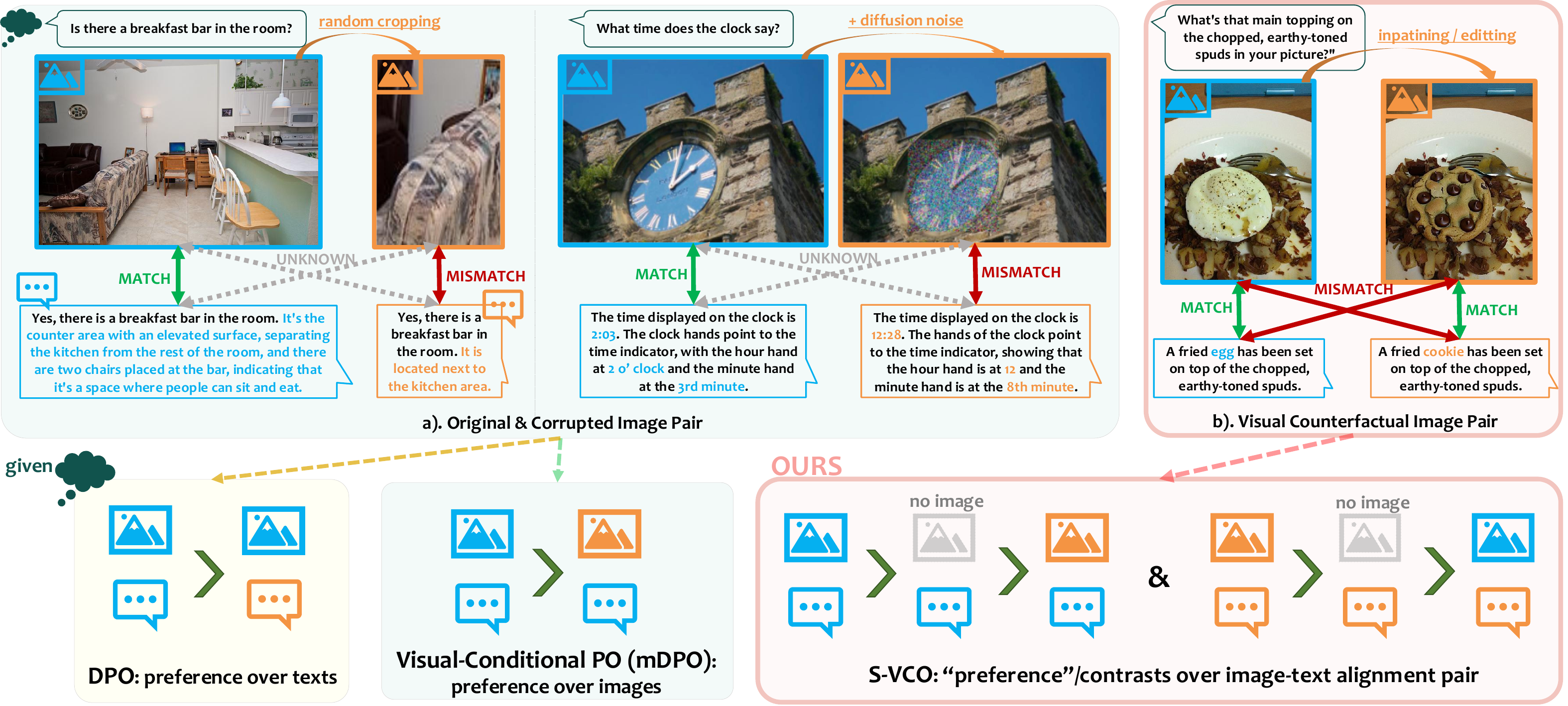}
\caption{\textbf{Upper Part: \dataname of visual counterfactual images [b)] in comparison to the image pair data used in prior work [a)] (\Cref{sec:data}).} \dataname's image pair differs in meaningful visual details that are also grounded in the associated texts [b)], while corrupting original images with random cropping or adding noise leads to images that are not aligned with the texts directly derived from language preference data [a)]. \textbf{Lower Part: \modelname in comparison to existing VLM preference tuning paradigms DPO and visual-conditional PO (\Cref{sec:model}).} Unlike prior methods that treat visual supervision as uni-modal preferences, \modelname considers the contrast of the image-text pair as a whole. It rewards the model for attending to matching images and rejecting contradictory ones (\Cref{sec:model_visual_contrastive}), while using a symmetrical mechanism to switch the role of each image-text pair, thus avoiding shortcut learning (\Cref{sec:model_symmetry}).}
\label{fig:data_and_objective_compare}
\end{figure*}

Recent works \citep{mdpo, vdpo, mfpo, imagedpo, chip} attempt to address similar issues of visual neglect by adapting Direct Preference Optimization (DPO)~\citep{dpo} to compare image inputs. 
However, these methods treat visual supervision as a ``preferential'' tuning paradigm, where an original image is preferred, and a cropped or noisy version of that image is dispreferred. 
As shown in \Cref{fig:data_and_objective_compare}, these noisy images lack meaningful connections to their paired texts, making it easy for the model to learn shortcuts by rejecting the corrupted image versions without fully understanding visual details and aligning them to the corresponding texts.

Instead, we posit that the ``\textbf{preference}'' paradigm should be improved with a purely \textbf{contrastive} framework, as the ``dispreferred'' image is merely another image misaligned with a given text. 
Building upon this insight, we propose \textbf{S}ymmetrical \textbf{V}isual \textbf{C}ontrastive \textbf{O}ptimization (\textbf{\modelname}), a finetuning objective that enforces precise correspondence between visual details and textual tokens.
\modelname rewards the model for attending to matching images and strongly rejecting contradictory images with incorrect details. 
To further avoid shortcut learning, \modelname incorporates \textbf{symmetry} by flipping the objective for contradictory responses, allowing the ``negative'' image to serve as the ``preferred'' visual condition when paired with its corresponding text.

Additionally, as cropping \cite{mdpo} or adding diffusion noise to an original image \cite{mfpo} fails to provide meaningful comparisons in visual details, we construct \textbf{\dataname}, a dataset of paired images with \textbf{M}inimal \textbf{V}isual \textbf{C}ontrasts, each matched to contrastive textual responses given a shared query. Building on recent visual counterfactual data sources \citep{countercurate, finecopsref}, we implement a vision-centric filter to select visually challenging pairs and an LLM augmentation scheme to diversify texts, forming an instruction-response-styled dataset suitable for VLM finetuning.

Experiments demonstrate the effectiveness and versatility of our approach, as \modelname consistently enhances VLM performance across diverse benchmarks spanning various abilities and domains. As shown in \Cref{fig:improve_per_domain}, \modelname achieves significantly greater improvements over the base-VLM, particularly in reducing visual hallucinations, while excelling in vision-centric tasks and offering considerable gains on general benchmarks. Compared to existing VLM preference tuning methods (DPO,~\citealp{dpo}; mDPO,~\citealp{mdpo}), \modelname combined with \dataname delivers more substantial and comprehensive performance boosts.

In summary, our main contributions are:

\textbullet \ \modelname, a novel VLM finetuning objective that enforces strict and balanced visual contrastive supervision through the symmetrical alignment of image-text pairs;

\textbullet \ \dataname, a dataset of minimal contrastive image pairs accompanied with corresponding textual responses for a shared query. \dataname is constructed automatically through vision-centric filtering and augmentation based on visual counterfactual data;

\textbullet \ The combination of \modelname and \dataname significantly boosts VLM performance across diverse benchmarks, particularly in visually dependent tasks, without compromising general capabilities.

\section{Preliminaries} \label{sec:preliminary}
Our visual contrastive objective is inspired by Direct Preference Optimization (DPO)~\citep{dpo} that contrasts pairs of textual responses. When applied to VLMs, DPO incorporates the image as an additional prefix condition \citep{povid, silkie, rlhfv, dpa}. Let $\pi_{\theta}$ be the policy VLM to be optimized, and $\pi_{\text {ref}}$ the fixed reference model (typically the unfinetuend VLM in DPO's framework) used to measure how the finetuned policy $\pi_{\theta}$ improves. Given a query $q$ (an instruction prompt), an image $i$, and a pair of responses -- one preferred $y_w$ (\textit{winning}) and one dispreferred $y_l$ (\textit{losing}) -- the \textbf{DPO} objective for VLMs can be formulated as:
\begin{equation}
\small
    L_{\text{DPO}} = -\log \sigma\left(\beta \log \tfrac{\pi_{\theta}\left(y_{w} \mid i, q\right)}{\pi_{\text {ref }}\left(y_{w} \mid i, q\right)}-\beta \log \tfrac{\pi_{\theta}\left(y_{l} \mid i, q\right)}{\pi_{\text {ref }}\left(y_{l} \mid i, q\right)}\right)
\label{eq:dpo}
\end{equation}

As illustrated in \Cref{fig:data_and_objective_compare}, the DPO formulation focuses on supervising differences between \textbf{language responses}, which however, often pertain to wording choices and stylistic variations rather than reflecting meaningful visual distinctions. Consequently, the VLM is being trained like a language model to weigh over text formulations, rather than to fully utilize the image as a grounding input.

Recent approaches, such as \citet{mdpo, mfpo}, adapt the DPO framework to contrast image conditions. The original image $i$ becomes the preferred image $i_w$ (\textit{winning}), while a negative image $i_l$ (\textit{losing}) is derived through random cropping \citep{mdpo} or adding diffusion noise \citep{mfpo} (see \Cref{fig:data_and_objective_compare}). The updated \textbf{Vis}ual \textbf{Con}ditional objective is:

{\small
\begin{align}
    &L_{\text{VisCon}} = \label{eq:visualdpo}\\ &-\log \sigma\left(\beta \log \tfrac{\pi_{\theta}\left(y_{w} \mid i_{w}, q\right)}{\pi_{\mathrm{ref}}\left(y_{w} \mid i_{w}, q\right)}-\beta \log \tfrac{\pi_{\theta}\left(y_{w} \mid i_{l}, q\right)}{\pi_{\mathrm{ref}}\left(y_{w} \mid i_{l}, q\right)}\right) \notag
\end{align}}

The above visual conditional preference formulation \textbf{shifts the target of preference onto the images}, where the original image $i_w$ is always preferred over the corrupted version $i_l$. However, this approach shares similar limitations with DPO, as it prioritizes distinguishing image differences without necessarily grounding those differences in the associated texts. Since $i_l$ is typically a noisy variant of $i_w$ with no definitive relations to $y_w$, the model could easily rely on superficial visual features to discern the image pair. This encourages shortcut learning, where the model rejects ``unrealistic'' images like $i_l$ without examining the visual details related to the text tokens.

Our \modelname addresses these limitations by introducing a stricter visual-conditioned objective (\Cref{sec:model_visual_contrastive}) and a symmetrical construct that aligns both $i_w$ and $i_l$ with their respective $y_w$ and $y_l$ (\Cref{sec:model_symmetry}). \Cref{fig:data_and_objective_compare} illustrates \modelname in comparison to existing preference tuning paradigms, showcasing its emphasis on grounded visual-textual alignment.

\section{\modelname: Symmetrical Visual Contrastive Optimization} \label{sec:model}

\subsection{Visual Contrastive Supervision} \label{sec:model_visual_contrastive}

\modelname enforces a strict visual focus by optimizing the model for two key behaviors:

\myparagraph{1) Attending to matching images} The model is rewarded for prioritizing relevant visual details in the matching image $i_w$ as a condition when predicting the corresponding response $y_w$. This directly addresses the tendency of VLMs to overlook visual content (\Cref{sec:intro}) and is achieved through the term:

{\small
\begin{align}
&L_{\text{Attend}}(i_w, y_w) =   \label{eq:attend} \\
&-\log \sigma\left(\beta_{1} \log \tfrac{\pi_{\theta}\left(y_{w} \mid i_{w}, q\right)}{\pi_{\mathrm{ref}}\left(y_{w} \mid i_{w}, q\right)}-\beta_{1} \log \tfrac{\pi_{\theta}\left(y_{w} \mid q\right)}{\pi_{\mathrm{ref}}\left(y_{w} \mid q\right)}\right). \notag
\end{align}}

\myparagraph{2) Rejecting contradictory images} When presented with a contrastive image $i_l$ containing visual details that directly contradict the response $y_w$, the model must strongly reduce the likelihood of predicting $y_w$ under this incorrect image condition. Intuitively, the model should assign minimal probability to a response that directly contrasts with the visual input. This behavior is modeled as:

{\small
\begin{align}
& L_{\text{Reject}}(i_l, y_w) = \label{eq:reject}
 \\
&-\log \sigma\left(\beta_{2} \log \tfrac{\pi_{\theta}\left(y_{w} \mid q\right)}{\pi_{\mathrm{ref}}\left(y_{w} \mid q\right)}-\beta_{2} \log \tfrac{\pi_{\theta}\left(y_{w} \mid i_{l}, q\right)}{\pi_{\mathrm{ref}}\left(y_{w} \mid i_{l}, q\right)}\right) \notag
\end{align}}

By combining these two components, our strict visual contrastive objective is defined as:
\begin{equation}
\small
    L_{\text{VCO}}(i_w, y_w, i_l) = L_{\text{Attend}}(i_w, y_w) + L_{\text{Reject}}(i_l, y_w)
\label{eq:vco}
\end{equation}

\subsection{Symmetrical Alignment} \label{sec:model_symmetry}

Unlike prior ``preference''-based approaches, \modelname treats $i_w$ and $i_l$ as mere images with contrastive details, where either can serve as the correct (\emph{i.e.}, ``preferred'') condition depending on the paired textual response ($y_w$ or $y_l$). While we inherit the notation of $i_w$ and $i_l$, \modelname does not assign an inherent ``winning'' or ``losing'' property to the images. Instead, an image is considered ``winning'' only when paired with its corresponding text. For instance, $i_l$, typically labeled as ``losing'' in preference tuning methods, becomes a ``winning'' (preferred) condition when the target response is $y_l$ that matches its visual details.

A one-sided formulation that consistently favors $i_w$ over $i_l$ risks encouraging shortcut learning, where the model rejects $i_l$ based on superficial, text-unrelated features (\emph{e.g.}, visual style differences). This issue could arise because most of the contrasting images are synthesized via inpainting or image editing (see \Cref{sec:data}). 
To address this, \modelname introduces symmetry by flipping the objective in \Cref{eq:vco}, treating $i_l$ as the preferred condition when paired with its corresponding response $y_l$. This effectively switches the roles of ``winning'' and ``losing'' for the image pair:
\begin{equation}
\small
    L_{\text{VCO}}(i_l, y_l, i_w) = L_{\text{Attend}}(i_l, y_l) + L_{\text{Reject}}(i_w, y_l).
\label{eq:vco_lose}
\end{equation}
As defined above, this encourages the model to attend to $i_l$ and reject $i_w$ in the flipped case where the target response is $y_l$.

The complete \modelname objective incorporates symmetry by summing over both roles of $i_w$ and $i_l$ given their corresponding target responses:
\begin{equation}
\small
    L_{\text{S-VCO}} = L_{\text{VCO}}(i_w, y_w, i_l) + L_{\text{VCO}}(i_l, y_l, i_w)
\label{eq:symmetric_vco}
\end{equation}

This symmetrical alignment ensures balanced optimization, allowing both $i_w$ and $i_l$ to contribute equally to the model’s learning. It promotes true alignment between images and texts without relying on shortcuts, such as rejecting either the image (\Cref{eq:visualdpo}) or the text (\Cref{eq:dpo}). In essence, \modelname optimizes for the alignment of \textbf{image-text pairs} rather than simply one modality.

\begin{figure*}[t]
\centering
\includegraphics[width=1.0\linewidth]{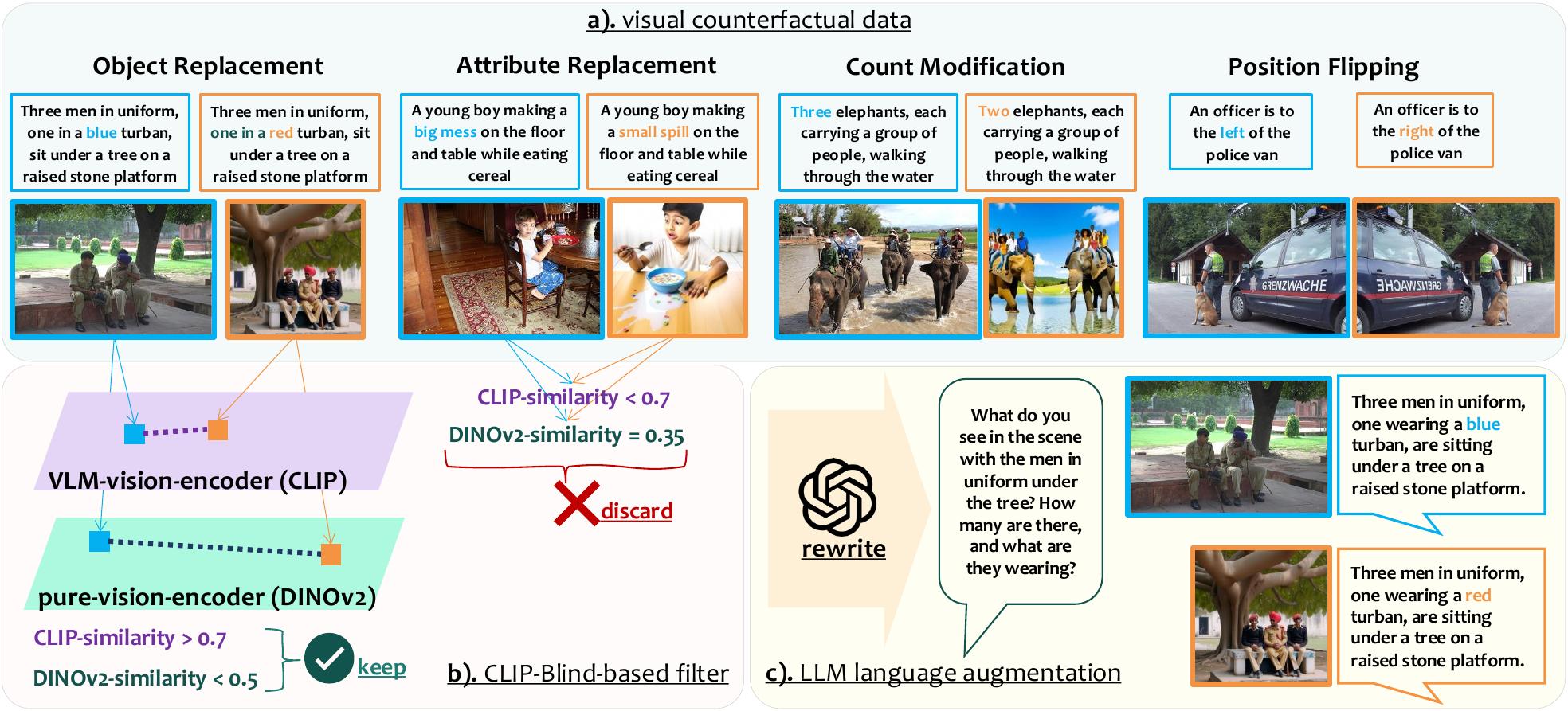}
\caption{\textbf{\dataname dataset outline}: 
\textbf{a).} Types of visual counterfactuals sourced from \citet{countercurate, finecopsref}; 
\textbf{b).} Our vision-centric filter that keeps only image pairs whose CLIP-similarity $> 0.7$ to select hard samples for current VLMs, while ensuring meaningful visual differences with DINOv2-similarity $< 0.5$; 
\textbf{c).} Rewriting captions into conversational queries and responses without changing the explicit minimal visual contrasts.}
\label{fig:data_piepline}
\end{figure*}

\section{Minimal Visual Contrasts Dataset} \label{sec:data}

\subsection{Visual Counterfactual Data} \label{sec:data_visual_counterfactual_data}
To complement \modelname, we introduce \textbf{\dataname}: a dataset of paired visual contrastive samples designed to enhance the model's ability to discern visual details. Built on existing sources (\emph{e.g.}, \citealp{countercurate, finecopsref, d3}), \dataname contains image pairs with minimal but meaningful variations, accompanied by corresponding contrastive texts.
The curated data includes four key contrast types, as shown in part \textbf{a).} of \Cref{fig:data_piepline}: 
\textbf{Object Replacement} (changing a specific object);
\textbf{Attribute Replacement} (modifying an object's features like color, shape, or size);
\textbf{Count Modification} (altering the number of objects); 
\textbf{Position Flipping} (reversing relative positions of objects). 
Except for the last type, these counterfactual images are generated through controlled image synthesis, including inpainting, editing, and generation \citep{gligen, powerpaint, dalle3}. In this work, we use CounterCurate \citep{countercurate} and FineCops-Ref \citep{finecopsref} as our visual counterfactual data sources. Refer to \Cref{sec:data_details} for dataset statistics.

\subsection{Filtering and Language Augmentation} \label{sec:data_filter_aug}
\myparagraph{Filter} 
Existing visual counterfactual datasets offer a large quantity of detailed contrasts, but their quality is inconsistent due to the synthetic nature of the data, which could lead to pairs where the generated image fails to truly contradict the original. 
To address this, we implement a vision-centric filter inspired by the CLIP-Blind concept \citep{mmvp} to select image pairs based on the following two criteria, as shown in part \textbf{b).} of \Cref{fig:data_piepline}:
\textbf{1. Different in detailed visual features}: Image pairs must exhibit meaningful contrasts, especially in detailed visual features. To achieve this, we employ DINOv2 \citep{dinov2}, a vision-only model with more robust visual feature representations \citep{MAEprepre}. Pairs with high similarity in DINOv2's purely visual representation space are discarded, as they may not contain the desired degree of contrasts.
\textbf{2. Semantically close \& Hard for VLMs}: Image pairs must also be semantically similar overall and difficult for current VLMs to distinguish. Therefore, we embed images using the same CLIP vision encoder used by the VLM \citep{clip, siglip} and retain pairs with relatively high similarity in the CLIP space that focuses on images' overall visual semantics. In this way, we exclude pairs with overly distinct content and less significant contrasts in visual details.
In practice, we use \texttt{DINOv2-Large} \citep{dinov2} with a similarity threshold of $0.5$, and \texttt{SigLIP-400M} \citep{siglip} as the CLIP encoder with a similarity threshold of $0.7$. Together, these thresholds ensure that \dataname focuses on meaningful visual contrasts while maintaining semantic relevance and difficulty for the VLM.

\myparagraph{Language Augmentation} 
Although visual counterfactual data sources provide image contrasts, their original textual descriptions are typically short captions, which are unideal for VLM finetuning. To address this, we augment the data using a two-step process with a strong LLM (\texttt{gpt-4o}), as shown in part \textbf{c).} of \Cref{fig:data_piepline}:
\textbf{1. Generating queries}: For each pair of captions, we prompt the LLM to generate a conversational question as if the captions were natural responses to that question, while ensuring that the contrasts remain explicit.
\textbf{2. Rewriting and diversifying}: We prompt the LLM to rephrase the original captions lacking clarity on key contrasts, thus enabling a more natural flow of language given the generated question and the contrastive details. 
Overall, the augmentation step results in conversational instruction-response pairs that are more suited for VLM finetuning, and more aligned with the visual details in the contrasting image pairs. Our prompt templates are provided in \Cref{sec:gpt4_prompt}.

After the filtering and language augmentation, our \dataname dataset comprises over $11,000$ pairs of minimal contrastive images matched with accurate and diverse conversational queries and responses (\Cref{sec:data_details}). The effectiveness of these data processing steps is empirically discussed in \Cref{sec:experiment_ablations}.

\begin{table*}[t]
\centering
\scriptsize
\begin{tabular}{llcccccccccc}
\toprule
\multicolumn{2}{r}{\textbf{Benchmarks}} & \multicolumn{2}{c}{\textbf{Hallucination}} & \multicolumn{3}{c}{\textbf{Vision-Centric}} & \multicolumn{2}{c}{\textbf{General}} & \multicolumn{1}{c}{\textbf{OCR}} & \multicolumn{1}{c}{\textbf{Knowledge}} & \multicolumn{1}{c}{\textbf{TOTAL}} \\ 
\cmidrule(lr){3-4} \cmidrule(lr){5-7} \cmidrule(lr){8-9} \cmidrule(lr){10-10} \cmidrule(lr){11-11} \cmidrule(lr){12-12}
& & \multicolumn{2}{c}{\textbf{MMHal}} & \textbf{CVBench} & \textbf{MMVP} & \textbf{RQA} & \textbf{MMVet} & \textbf{LVBench} & \textbf{TextVQA} & \textbf{SQA}& \textbf{avg\_impr.}\\ 
\multicolumn{2}{c}{\textbf{Models}$_{\text{\textbf{TrainSet}}}$}& \textbf{score}& \textbf{hal\_rate$\downarrow$} & \textbf{acc.} & \textbf{acc.} & \textbf{acc.} & \textbf{score} & \textbf{score} & \textbf{acc.} & \textbf{acc.} & \textbf{over BASE}\\
\midrule
\multirow{6}{*}{\rotatebox{90}{\textbf{LV-1.5-7B}}}
& BASE                  & $2.16$    & $57\%$    & $59.3$    & $21.3$& $35.3$    & $30.46$    & $61.2$    & $46.40$    & \underline{$66.78$}    & $0\%$ \\
& DPO$_{\text{VLF}}$    & $2.06$    & $65\%$    & $57.0$    & $16.7$& $39.5$    & $31.65$    & $68.1$    & \underline{$49.16$}    & $66.71$    & $-1.25\%$\\
& DPO$_{\text{MVC}}$    & \underline{$2.45$}    & \underline{$53\%$}    & \underline{$63.2$}    & \underline{$22.0$}& $42.1$    & \underline{$33.53$}    & $66.0$    & $\mathbf{49.43}$    & $66.07$    & \underline{$+8.11\%$}\\
& mDPO$_{\text{VLF}}$   & $2.39$    & $57\%$    & $53.2$    & $18.7$& $\mathbf{44.2}$    & $31.79$    & \underline{$68.4$}    & $41.71$    & $66.52$    & $+2.11\%$\\
& mDPO$_{\text{MVC}}$   & $2.29$    & $56\%$    & $59.4$    & $20.7$& $35.2$    & $31.51$    & $63.0$    & $46.42$    & $66.80$    & $+1.26\%$\\
& S-VCO$_{\text{MVC}}$ & $\mathbf{2.75}$ & $\mathbf{46\%}$    & $\mathbf{63.5}$    & $\mathbf{25.3}$& \underline{$43.0$}    & $\mathbf{34.68}$    & $\mathbf{69.5}$    & \underline{$49.16$}    & $\mathbf{67.22}$    & $\mathbf{+14.26\%}$\\
\midrule
\multirow{9}{*}{\rotatebox{90}{\textbf{LV-INT-7B}}}
& BASE                  & $2.74$           & $46\%$             & $67.6$         & $41.3$& $57.5$         & $41.01$         & $74.5$         & $59.45$         & \underline{$74.77$}          & $0\%$ \\
& DPO$_{\text{VLF}}$    & $2.70$           & $48\%$             & $68.1$         & $42.7$& $54.4$         & $40.87$         & $81.0$         & $59.20$         & $\mathbf{74.79}$          & $+0.10\%$\\
& DPO$_{\text{MVC}}$    & $2.92$           & \underline{$38\%$}    & $68.8$         & \underline{$46.7$}& \underline{$57.9$}     & \underline{$41.51$} & \underline{$83.5$}         & \underline{$59.76$} & $73.80$ & \underline{$+5.78\%$}\\
& mDPO$_{\text{VLF}}$   & $2.97$ & $42\%$           & $68.0$         & $42.0$& $54.8$         & $40.46$         & $79.3$ & $58.97$         & $74.53$          & $+2.07\%$\\
& mDPO$_{\text{MVC}}$   & \underline{$3.04$}           & \underline{$38\%$} & \underline{$70.0$} & $44.7$& \underline{$57.9$} & $41.24$         & $80.2$         & $59.43$         & $74.65$ & $+5.43\%$\\
& S-VCO$_{\text{MVC}}$ & $\mathbf{3.28}$ & $\mathbf{35\%}$    & $\mathbf{71.0}$ & $\mathbf{50.7}$& $\mathbf{58.2}$ & $\mathbf{44.04}$ & $\mathbf{85.0}$ & $\mathbf{60.26}$ & $73.87$          & $\mathbf{+10.47\%}$\\
\cmidrule(lr){2-12}
& S-VCO$_{\text{MVC-Raw}}$                  & $3.10$           & $35\%$             & $71.3$ & $46.0$& $58.8$        & $40.55$        & $86.5$ & $59.22$         & $73.87$       & $+7.73\%$\\
& VCO$_{\text{MVC}}$    & $3.16$           & $39\%$             & $70.1$         & $46.0$& $56.2$         & $42.61$         & $84.8$         & $59.87$         & $74.44$ & $+6.82\%$\\
& SFT$_{\text{MVC}\times2}$    & $2.68$           & $46\%$    & $66.6$         & $38.7$ & $56.5$     & $38.94$ & $70.7$         & $59.32$ & $74.77$ & $-2.45\%$\\
\bottomrule
\end{tabular}
\caption{\textbf{Performance of different methods applied to two base-VLMs, tested across benchmarks grouped by ability domains}. VLF refers to VLFeedback used in mDPO; MVC is our minimal visual contrastive dataset; RQA and SQA represent RealworldQA and ScienceQA (\Cref{sec:experiment_setup}). The last column shows the average percentage of improvement over the base-VLM across all metrics. \textbf{Best scores are in boldface}, \underline{second-best underlined}. 
Our \modelname demonstrates consistent improvement across domains, achieving the most significant enhancement over the base-VLMs overall. 
The last three rows present ablation results (\Cref{sec:experiment_ablations}): 
1. \modelname on unfiltered and unaugmented visual counterfactual data (MVC-Raw); 
2. \modelname without the symmetrical construct (VCO, \Cref{sec:model_visual_contrastive}); 
3. SFT using both sides of image-text pairs from \dataname (SFT$_{\text{MVC}\times2}$). 
These results highlight the importance of our data preprocessing (\Cref{sec:data_filter_aug}) and the symmetrical objective (\Cref{sec:model_symmetry}) for optimal performance.}
\label{tab:benchmark_results_all}
\end{table*}

\section{Experiments} \label{sec:experiment}
\subsection{Setup} \label{sec:experiment_setup}

\myparagraph{Baseline methods} 
We compare \modelname against DPO~\citep{dpo} and mDPO~\citep{mdpo}\footnote{In practice, mDPO combines textual DPO (\Cref{eq:dpo}) with the visual-conditional version (\Cref{eq:visualdpo}) and an absolute reward regularization.} containing the visual-conditional objective in \Cref{eq:visualdpo} as baseline finetuning approaches.

\myparagraph{Training data} 
We use two datasets: our proposed \dataname with minimal contrastive image-text pairs and VLFeedback (\textsc{VLF})~\citep{silkie}, a classical instruction-tuning dataset for VLMs that includes preferred and dispreferred response pairs. We follow mDPO~\citep{mdpo}'s sample of \textasciitilde10,000 data points from \textsc{VLF}. Refer to \Cref{sec:data_details} for more dataset implementation details.

\myparagraph{Base VLMs} 
We use two pretrained models hosted on \href{https://huggingface.co/}{Huggingface}: \href{https://huggingface.co/llava-hf/llava-1.5-7b-hf}{\texttt{LLaVA-1.5-7B}}~\citep{llava1.5} -- denoted as \textsc{LV-1.5(-7B)}, and \href{https://huggingface.co/llava-hf/llava-interleave-qwen-7b-hf}{\texttt{LLaVA-Next-Interleave-7B}}~\citep{llavanextinterleave} -- denoted as \textsc{LV-INT(-7B)}.

\myparagraph{Evaluation} 
We evaluate the models on a wide range of benchmarks spanning various ability domains. Following the categorization by \citet{cambrian}, these benchmarks include: 
\textit{General}: LLaVABench~\citep{llava}, MMVet~\citep{mmvet};
\textit{Hallucination}: MM-Hal~\citep{sun2023aligning}; 
\textit{Vision-Centric}: CVBench~\citep{cambrian}, MMVP~\citep{mmvp}, RealworldQA~\citep{realworldqa};
\textit{OCR}: TextVQA~\citep{textvqa};
\textit{Knowledge}: SQA~\citep{sqa}.

\myparagraph{Implementation details} Refer to \Cref{sec:implement_details} for detailed training and inference configurations.

\begin{figure*}[!t]
\centering
\includegraphics[width=1.0\linewidth]{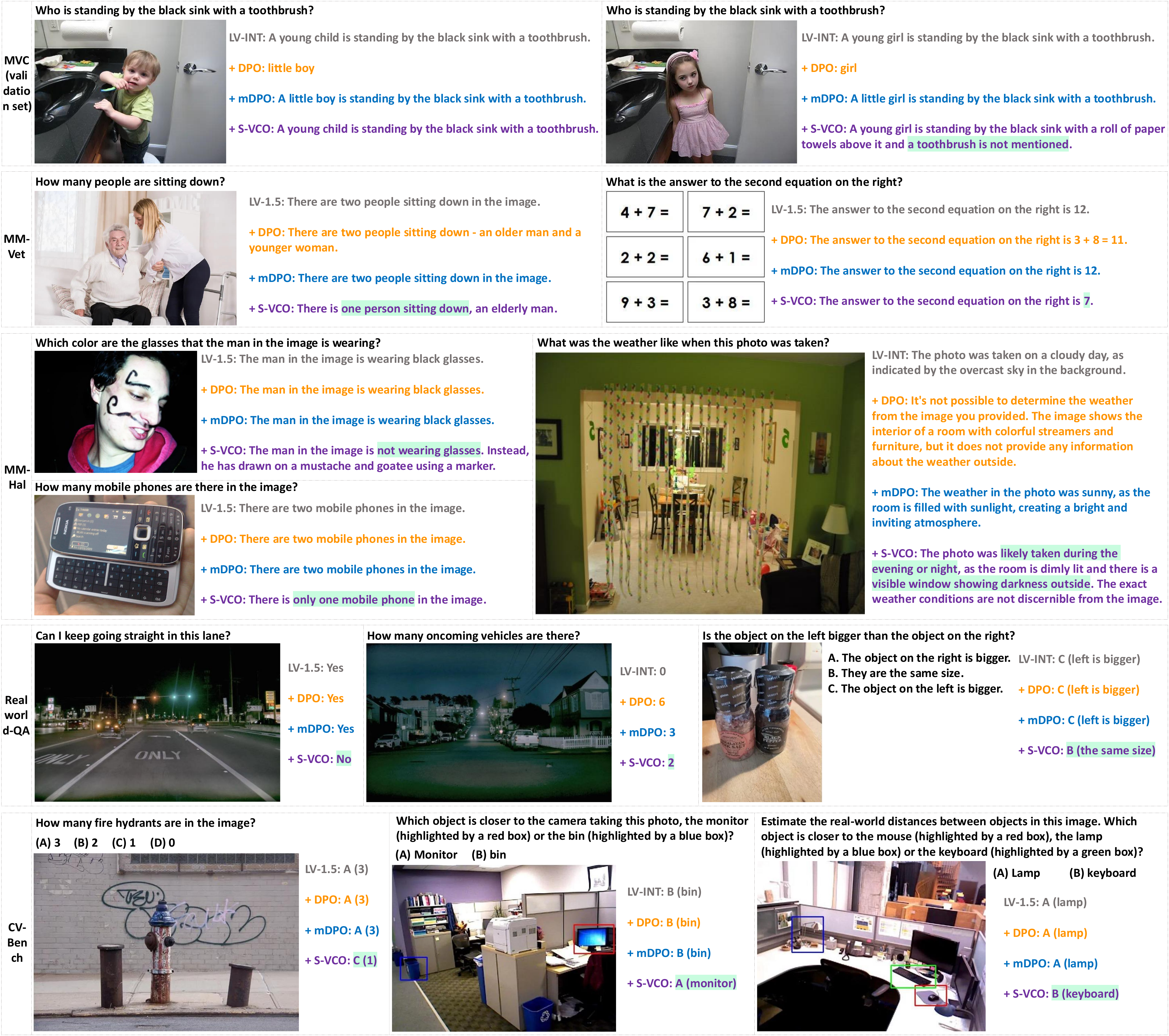}
\caption{\textbf{Qualitative examples extracted from various benchmarks} comparing \textcolor{baseColor}{\textbf{base-VLMs} (\textsc{\textbf{LV-INT}} or \textsc{\textbf{LV-1.5}})} to the results after finetuning with \textcolor{dpoColor}{\textbf{DPO}}, \textcolor{mdpoColor}{\textbf{mDPO}} or \textcolor{svcoColor}{\textbf{\modelname}} on \dataname dataset. \colorbox{goodShadeColor}{Accurate captions of visual information are highlighted}. 
Our method \textcolor{svcoColor}{\modelname} demonstrates superior understanding of \textbf{fine-grained visual details} (\emph{e.g.}, identifying the absence of a toothbrush) and shows \textbf{strong resilience to visual hallucinations} (\emph{e.g.}, recognizing marker-drawings, fire hydrants, slide-phones). Furthermore, \modelname excels in more advanced \textbf{visual reasoning} (\emph{e.g.}, interpreting drive-lane conditions \& regulations, estimating object sizes \& distances), and captures \textbf{complex scenes} with greater detailedness and depth (\emph{e.g.}, identifying weather through the window, recognizing oncoming vehicles in low-light settings).}
\label{fig:qual_samples}
\end{figure*}

\begin{figure}[t]
\centering
\includegraphics[width=1.0\linewidth]{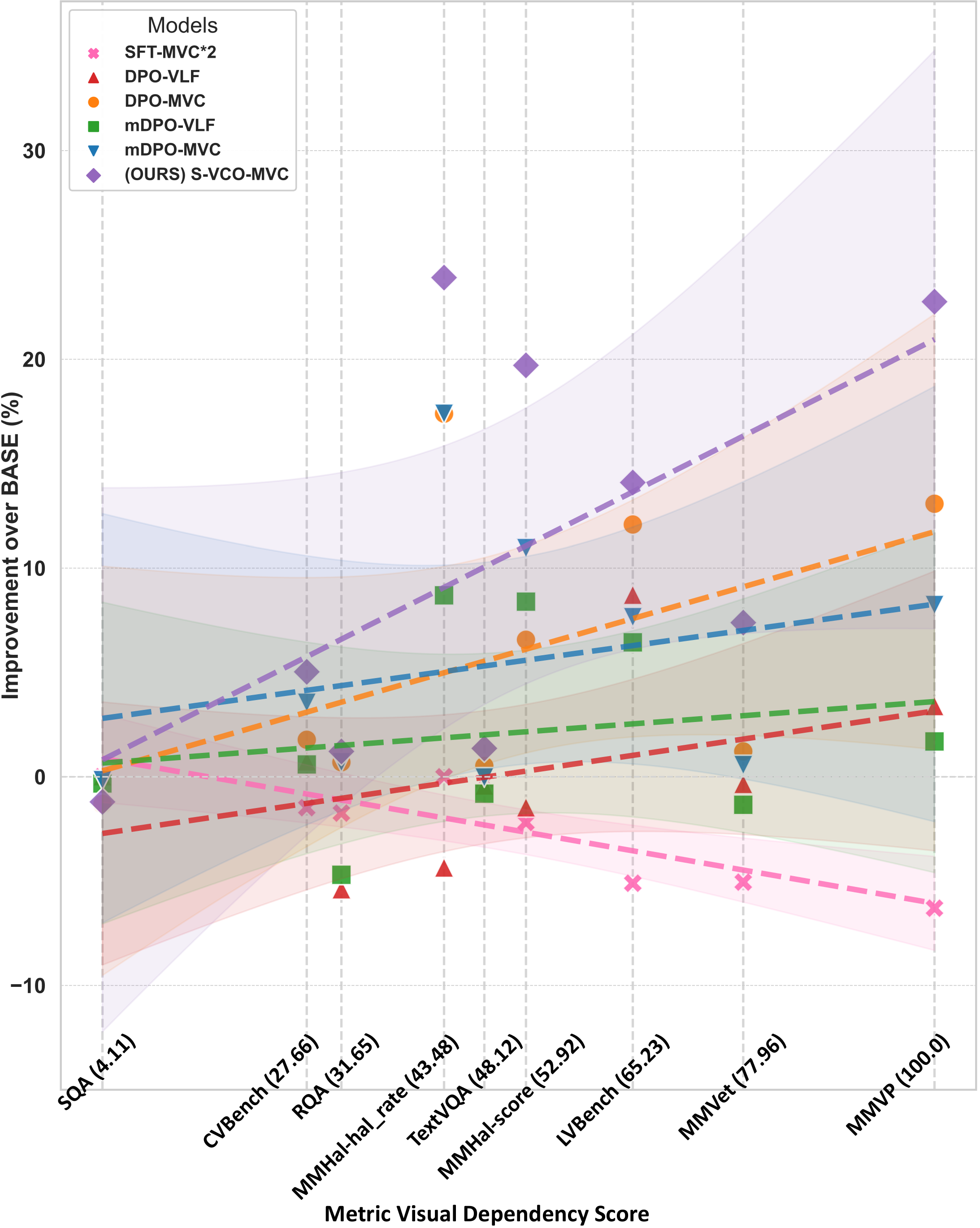}
\caption{\textbf{Trend of improvement over the base-VLM as benchmarks become increasingly visually dependent.} A metric's visual dependency is measured as the performance drop of the base-VLM when no image input is provided. 
\modelname exhibits the most significant trend of improvements with increasing task visual dependency, highlighting how its objective design (\Cref{sec:model}) enhances model's focus on critical visual details.
\dataname dataset also strengthens existing preference tuning methods (DPO and mDPO), while SFT (\Cref{sec:experiment_ablations}) degrades performance on more visually demanding benchmarks.}
\label{fig:improve_per_visual_dependency}
\end{figure}

\subsection{Main Results} \label{sec:experiment_results}
\Cref{tab:benchmark_results_all} presents detailed evaluation results for different methods across various benchmarks grouped by ability domains. The final column shows the average improvement (\%) over the base VLM across all metrics, summarizing each model's overall performance. Key findings are highlighted below.

\myparagraph{\modelname achieves consistently superior performance across benchmarks in various domains} 
\Cref{fig:improve_per_domain} illustrates performance improvements over the base-VLM (\textsc{LV-INT}) for each benchmark category. Our \modelname consistently outperforms baseline methods across nearly all domains, with only a slight drop in knowledge-heavy tasks like ScienceQA that relies minimally on visual input (discussed more below). 
The most significant gain is in visual hallucination tasks, where \modelname enhances the base model by \textbf{\textasciitilde$\mathbf{22}$\%}. 
Considerable improvements are also observed in vision-centric (\textbf{$\mathbf{+}$\textasciitilde$\mathbf{10}$\%}) and general domains (\textbf{$\mathbf{+}$\textasciitilde$\mathbf{11}$\%}).
Compared to baseline methods, DPO and mDPO trained on the standard instruction-tuning dataset \textsc{VLF}, \modelname with \dataname delivers much greater and more consistent improvements across domains. 
While recent approaches like mDPO~\citep{mdpo} improve visual hallucination metrics, its effects on other abilities are less notable. 
In contrast, \modelname not only excels in visually demanding tasks but also achieves considerable gains in other domains.

\myparagraph{\modelname shows increasing benefits as benchmarks become more visually dependent} 
We quantify \textbf{visual dependency} of a metric as the percentage drop in a base-VLM's performance when image inputs are removed. 
Using \textsc{LV-INT} as the base model, we rank benchmarks by visual dependency (\emph{e.g.}, MMVP accuracy drops to $0$ without image inputs), and plot the improvement trends of different methods in \Cref{fig:improve_per_visual_dependency}. Dotted lines show fitted regressions, and shaded areas represent variances.
\modelname demonstrates increasingly pronounced improvements as visual dependency rises, aligning with its design focus on strengthening visual detail recognition (\Cref{sec:model}). 
On highly visually-dependent benchmarks such as MM-Hal, LLaVABench, MMVet and MMVP, our \modelname delivers the most substantial gains. 
In contrast, on SienceQA (SQA), which benefits merely \textbf{\textasciitilde$\mathbf{4}$\%} from image inputs, we observe a minor drop in performance after \modelname, as visual information plays very little role in this task.
Compared to other methods, SFT degrades performance on visually dependent metrics (\Cref{sec:experiment_ablations}), while preference tuning approaches show positive trends. 
Among all, \modelname achieves the most significant trend of gains with increasingly vision-intensive tasks.

\myparagraph{Qualitative performance highlights \modelname's superior visual understanding} 
In \Cref{fig:qual_samples}, qualitative examples from various benchmarks further demonstrate \modelname's superior ability to process fine-grained visual details and reason about complex scenes. 
\modelname excels in recognizing subtle yet critical visual distinctions (\emph{e.g.}, identifying the absence of a toothbrush) and remains robust against visual hallucinations (\emph{e.g.}, differentiating marker-drawings, slide-phones, and fire hydrants). 
Moreover, \modelname shows strong visual reasoning capabilities by interpreting complex scenarios such as drive-lane conditions. It also captures intricate details in scenes with greater depth and contextual awareness (\emph{e.g.}, discerning weather conditions through a window or identifying oncoming vehicles in low-light settings).

\myparagraph{\dataname enhances previous preference tuning methods} 
Beyond its strong synergy with \modelname, \dataname dataset also strengthens the effects of existing preference tuning methods, particularly DPO. 
For \textsc{LV-INT}, both DPO and mDPO achieve greater overall improvements when trained on \dataname compared to the textual-instruction-tuning-styled dataset \textsc{VLF}. For \textsc{LV-1.5}, DPO$_{\text{MVC}}$ outperforms DPO$_{\text{VLF}}$ by \textbf{\textasciitilde$\mathbf{9}$\%} on average across benchmarks, achieving second-best results on multiple metrics behind only our \modelname.
\Cref{fig:improve_per_visual_dependency} further illustrates DPO and mDPO variants trained on \dataname outperforming their \textsc{VLF}-trained counterparts, especially on visually dependent tasks. By providing visually challenging image pairs matched with texts, \dataname proves to be a more effective training resource across preference tuning methods.

\subsection{Ablations} \label{sec:experiment_ablations}

\myparagraph{Data filter and augmentation} 
We evaluate the impact of our filtering and augmentation step (\Cref{sec:data_filter_aug}) in \dataname by training \textsc{LV-INT-7B} on unprocessed (``raw'') visual counterfactual data directly sourced from CounterCurate \citep{countercurate} and FineCops-Ref \citep{finecopsref}. 
We sample $11,149$ random datapoints exactly matching \dataname's size (\Cref{sec:data_details}), and train the model with \modelname using the same configurations.
The resulting model, \textbf{S-VCO$_{\text{MVC-Raw}}$}, underperforms S-VCO$_{\text{MVC}}$ trained on filtered and augmented data, with drops across most benchmarks (\Cref{tab:benchmark_results_all}). This highlights the importance of our filter and augmentation step in constructing high-quality contrastive data that better supports \modelname.
Notably, despite the data preprocessing omissions, S-VCO$_{\text{MVC-Raw}}$ still surpasses all other baseline preference tuning methods, demonstrating the robustness of \modelname's symmetrical visual contrastive objective.

\myparagraph{Symmetrical loss construction} 
A key feature of \modelname is its symmetrical loss, which treats both sides of the texts $y_w$ and $y_l$ as preferred when aligned with their respective images $i_w$ and $i_l$, optimizing both simultaneously (\Cref{sec:model_symmetry}). 
To assess the necessity of this symmetry, we train \textsc{LV-INT-7B} on \dataname without the symmetrical term $L_{\text{VCO}}(i_l, y_l, i_w)$, reducing the objective to a one-sided preference tuning approach. 
The resulting model, \textbf{VCO$_{\text{MVC}}$}, underperforms S-VCO$_{\text{MVC}}$, as shown in \Cref{tab:benchmark_results_all}, though it still surpasses all other baseline methods due to VCO's strong visual contrastive supervision (\Cref{sec:model_visual_contrastive}).
The performance drop is most evident in hallucination and vision-centric tasks, highlighting the importance of optimizing both sides of the contrastive pair. 
Symmetry mitigates shortcut learning by encouraging the model to focus on meaningful image-text alignments rather than superficial image features.

\myparagraph{Comparing to SFT} 
To investigate whether standard supervised finetuning (SFT) on both sides of the image-text pairs could achieve similar results, we construct a new instruction-tuning dataset by including both $(i_w, y_w)$ and $(i_l, y_l)$ from \dataname, effectively doubling its size (${ \text{MVC}\times2}$). 
Results for \textbf{SFT$_{\text{MVC}\mathbf{\times2}}$} are shown in the last row of \Cref{tab:benchmark_results_all}.
While SFT preserves performance on ScienceQA -- a benchmark minimally reliant on visual inputs -- it performs significantly worse across all other domains, yielding the lowest results among all methods. 
Unlike \modelname, SFT lacks the contrastive supervision necessary to highlight subtle visual-text alignment, thus leading to poor performance on tasks requiring strong visual grounding.

\section{Related Work} \label{sec:related}
\myparagraph{VLM's Visual Hallucinations}
Recent studies~\citep{deng2024seeing,mmvp,chen2024quantifying,symdpo} have shown that VLMs tend to hallucinate content not present in the visual input. VLMs also struggle with fine-grained visual understanding (\emph{e.g.}, recognizing object attributes and relations) -- especially when tested on confounding image pairs~\citep{peng2024synthesize,li2024naturalbench}. This aligns with our findings in~\Cref{fig:visual_neglect}, suggesting a lack of robust and faithful multimodal grounding.
To address this issue, several training-free methods have been proposed. \citet{yang2023set} introduced ``Set-of-Mark'' prompting that overlays spatial and textual markers on images to help models reference specific regions. \citet{deng2024seeing} employed CLIP-guided decoding to steer the language outputs with grounded visual cues. 
Architecture-wise, GRILL~\cite{jin2023grill} incorporates object-level alignment during pretraining to promote visual grounding.
Unlike previous approaches, our work focuses on finetuning with a novel objective (\Cref{sec:model}) and a data construction pipeline (\Cref{sec:data}) based on visual counterfactuals~\citep{countercurate, finecopsref}, targeting more precise alignment of multimodal details.

\myparagraph{VLM Finetuning} 
Finetuning improves task-specific performance of VLMs and aligns them better with human preferences.
SFT remains widely adopted to guide models toward towards following instructions~\citep{sun2024layoutvlm,jiang2024supervised}. 
DPO~\citep{dpo} optimizes the margin between finetuned and unfinetuned model versions using paired preference data. Its extension to VLMs incorporates image as additional prefix condition~\citep{povid}. 
Recent methods such as mDPO~\citep{mdpo}, MFPO~\citep{mfpo}, V-DPO~\citep{vdpo}, CHiP~\citep{chip} and Image-DPO~\citep{imagedpo} further adapt the preference tuning paradigm to focus on image-side preferences over a pair of ``good'' and ``bad'' image, aiming to reduce visual hallucinations.
Our approach \modelname replaces the one-sided ``preference'' formulation with a stricter visual contrastive objective of symmetrical construct, treating ``preference'' explicitly as alignment over matching image-text pairs. This enables more comprehensive and robust VLM improvements across tasks.

\section{Conclusion} \label{sec:conclusion}

This work introduces \modelname, a novel VLM finetuning objective that enforces strict visual contrastive supervision within a symmetrical construct. 
Complementing this objective, we propose \dataname, a dataset of paired images with minimal visual contrasts, each associated with corresponding contrastive texts. 
Experiments demonstrate that combining \modelname with \dataname consistently improves VLM performance across diverse benchmarks, with particularly significant gains on visually dependent tasks. Importantly, these improvements are achieved without compromising, and even enhancing VLMs' general capabilities. 

\section*{Limitations} \label{sec:limitations}

Our method incorporates multiple individual loss terms and weights (\Cref{sec:model}), which may require manual tuning to determine the optimal settings. Adjusting these hyperparameters could potentially further enhance performance.

The \modelname objective works most effectively when the contrastive image pairs have meaningful differences in visual details. The current \dataname derived from existing visual counterfactual data sources includes a limited set of operations for building the contrasts (\Cref{sec:data_visual_counterfactual_data}). Our method could benefit from a more diverse set of contrasting visual details that should enable the model to potentially learn a broader range of visual features.

\section*{Ethics Statement} \label{sec:ethics}
In this work, all data and pretrained models are publicly available. They are collected and processed in adherence to the respective data, checkpoints, and API usage policy. We acknowledge that our finetuned models may generate unsafe content, and we advise all users of careful verification before deploying this work in real-world applications.

\bibliography{main}

\newpage
\appendix

\section{Dataset Details} \label{sec:data_details}

\Cref{tab:counterfactual_data_stats} details the dataset statistics of our visual counterfactual data sources CounterCurate~\citep{countercurate} and FineCops-Ref~\citep{finecopsref}, as well as our \dataname after filtering and augmentation (disccused in \Cref{sec:data_filter_aug}). 
We discard the original ``Order'' category of FineCops-Ref for the frequent irrational cases under that category. For the non-synthetic ``Left-Right'' position flipping, we did not apply the filter but randomly sampled the data proportional to its original category distribution in the source datasets. 

When training with \dataname, we leave out $240$ random samples for validation set ($200$ from CounterCurate and $40$ from FineCops-Ref). This leads to a total training data size of $10,909$ with \dataname. 
When training with the sampled VLFeedback data \citep{silkie}, used as in mDPO~\citep{mdpo} (\url{https://huggingface.co/datasets/fwnlp/mDPO-preference-data}; noted as \textsc{VLF}), we leave out $200$ random samples for validation set, leading to a total training data size of $9,222$ with \textsc{VLF}.

\begin{table}[ht]
\centering
\scriptsize
\resizebox{\columnwidth}{!}{%
\begin{tabular}{lccc}
\toprule
 & \textbf{CounterCurate} & \textbf{FineCops-Ref} & \textbf{\dataname} \\
\midrule
Object Replacement    & $26,164$ & $4,171$ & \multirow{2}{*}{$7,189$} \\
Attribute Replacement & $27,964$ & $1,844$ &  \\
Count Modification    & $10,010$ &    $0$ & $919$ \\
Position Change       & $56,711$ & $1,555$ & $3,041$ \\
Total                & $120,849$ & $7,570$ & $11,149$ \\
\bottomrule
\end{tabular}
}
\caption{\textbf{Statistics of visual counterfactual datasets} CounterCurate, FineCops-Ref, and our \dataname after filtering and augmenting both data sources (\Cref{sec:data}). For \dataname, the number of samples in the categories ``Object Replacement'' and ``Attribute Replacement'' are counted together.}
\label{tab:counterfactual_data_stats}
\end{table}

\section{Implementation Details} \label{sec:implement_details}

\myparagraph{Training} 
We set $\beta_{1}$ and $\beta_{2}$ in \modelname's objective (\Cref{eq:symmetric_vco}) to $0.1$, following the typical $\beta$ values in DPO (\Cref{eq:dpo}) and mDPO (\Cref{eq:visualdpo}). 
All models are finetuned for $2$ epochs with a batch size of $32$ on $8$ \textsc{NVIDIA-A100x80G} GPUs.
When training on \textsc{VLF}, we retain the original learning rate of $1e{-05}$ used by DPO and mDPO. The text sequence length during finetuning is set to $1024$ to accommodate \textsc{VLF}'s text data length.
When training on our \dataname, we set a learning rate of $1e{-06}$ for all methods except mDPO on \textsc{LV-1.5-7B}, where a lower rate of $1e{-07}$ is used to better stabilize training. The text sequence length during finetuning is set to $128$ given the \textsc{MVC}'s shorter text data length.
Intermediate checkpoints are saved at an interval of $24$, $31$, and $62$ steps for training on \textsc{VLF}, \dataname, and \textsc{MVC}$\times2$ (used for SFT ablation, see \Cref{sec:experiment_ablations}), respectively.

\myparagraph{Evaluation} 
We report results using the best-performing checkpoint for each model configuration (\Cref{tab:benchmark_results_all}), selected based on the highest average improvement over the base-VLM. 
For all benchmarks, the temperature of model predictions is set to $0$. 
As the evaluation judge, we use \texttt{gpt-4-0613} for LLaVABench~\citep{llava} and MMVet~\citep{mmvet}, and \texttt{gpt-4-turbo} for MM-Hal~\citep{sun2023aligning}.

\section{Prompt for \dataname Language Augmentation} \label{sec:gpt4_prompt}

Below are our prompt templates for querying GPT4 (\texttt{gpt-4o}) to augment the queries and responses from visual counterfactual data sources (\Cref{sec:data_filter_aug}). 

In the first step, we ask GPT4 to generate an natural and appropriate question that targets the subtle differences in the response-pair.

In the second step, we ask GPT4 to revise the instruction generated from the first step, and rephrase the original response pairs to make the whole instruction-response conversation sound more natural, while retaining the contrastive details in the responses.

In both steps, we set the temperature to $0.7$ for more diversified wording.

\begin{tcolorbox}[float=t,
                  colback=black!5!white,
                  colframe=black!75!black,
                  title=GPT4 Language Augmentation Step 1]
\texttt{{\color{blue}\lbrack\text{System}\rbrack}}\\
You are a helpful assistant that generates a natural-sounding instruction prompt for a vision-language scenario. Given two responses about an image: one `chosen' and one `rejected', your task is to produce a single instruction or question that encourages the user to naturally reveal the critical differences between the two responses. Focus on attributes that differ (like number, color, position, orientation). The prompt should sound like a normal request someone might ask when wanting more detail about the image. It should not sound overly forced or contrived, and it should not explicitly mention that there are two responses or that differences are being tested. Also, try to vary your phrasing, and do not always start the instruction with `Could' or `Can'. \\
\\
\texttt{{\color{blue}\lbrack\text{User}\rbrack}}\\
Chosen response: \texttt{{\color{red}\{\text{ORIGINAL\_RESPONSE}\}}}\\
Rejected response: \texttt{{\color{red}\{\text{CONTRAST\_RESPONSE}\}}}\\
\\
Generate a single, natural-sounding instruction or question that would prompt a user or model to include the detail of the collar in a natural way. Avoid making the prompt sound forced or unnatural, and do not explicitly mention comparing two descriptions.
\end{tcolorbox}

\begin{tcolorbox}[float=t,
                  width=\textwidth,
                  colback=black!5!white,
                  colframe=black!75!black,
                  title=GPT4 Language Augmentation Step 2]
\small
\texttt{{\color{blue}\lbrack\text{System}\rbrack}}\\
You are a helpful assistant that revises instruction prompts and their corresponding responses for vision-language scenarios. Given an initial instruction and two responses about an image: one `chosen' and one `rejected', your task is to: \\
1. Revise the instruction to make it sound more natural and conversational and ensure it seamlessly leads to the given responses. Also, try to vary your phrasing, and do not always start the instruction with `Could' or `Can'.\\
2. Rephrase both the chosen and rejected responses to diversify the language without adding or removing any details.\\
3. Ensure that the differences between the chosen and rejected responses remain highlighted and are consistent with the original responses.\\
\\
Do not introduce any new information or omit existing details. The revisions should maintain accuracy and ensure coherence between the instruction and responses. \\
\\
\texttt{{\color{blue}\lbrack\text{User}\rbrack}}\\
Initial Instruction: \texttt{{\color{red}\{\text{STEP1\_GENERATED}\}}}\\
Chosen Response: \texttt{{\color{red}\{\text{ORIGINAL\_RESPONSE}\}}}\\
Rejected Response: \texttt{{\color{red}\{\text{CONTRAST\_RESPONSE}\}}}\\
\\
Revise the instruction and both responses as described above. \\
\\
Here are some examples: \\
\\
1.\\
Initial Instruction: "What can you tell me about the hair color of the woman who is sweeping the floor?"\\
Chosen Response: "A blonde-haired woman wearing a white skirt, white shirt, white apron, and black shoes is sweeping the floor."\\
Rejected Response: "A black-haired woman wearing a white skirt, white shirt, white apron, and black shoes is sweeping the floor."\\
Revised Instruction: "Describe the woman's hair color and her attire while she's sweeping the floor?"\\
Revised Chosen Response: "The woman sweeping the floor has blonde hair and is wearing a white skirt, white shirt, white apron, and black shoes."\\
Revised Rejected Response: "The woman sweeping the floor has black hair and is wearing a white skirt, white shirt, white apron, and black shoes."\\
\\
2.\\
Initial Instruction: "Where is the air stunt relative to the snowy mound in the image?"\\
Chosen Response: "An air stunt is above a snowy mound."\\
Rejected Response: "An air stunt is below a snowy mound."\\
Revised Instruction: "What do you notice about the position of the air stunt in relation to the snowy mound in the image?"\\
Revised Chosen Response: "The air stunt is positioned above the snowy mound."\\
Revised Rejected Response: "The air stunt is located below the snowy mound."\\
\\
Now, revise the following instruction and responses:\\
Initial Instruction: \texttt{{\color{red}\{\text{STEP1\_GENERATED}\}}}\\
Chosen Response: \texttt{{\color{red}\{\text{ORIGINAL\_RESPONSE}\}}}\\
Rejected Response: \texttt{{\color{red}\{\text{CONTRAST\_RESPONSE}\}}}\\
\\
Reply ONLY in the following format and no other text or notes:\\
\\
Revised Instruction: \\
Revised Chosen Response: \\
Revised Rejected Response: 
\end{tcolorbox}

\end{document}